# A Classical Approach to Handcrafted Feature Extraction Techniques for Bangla Handwritten Digit Recognition

Md. Ferdous Wahid
*Department of Electrical and Electronic Engineering*
*Hajee Mohammad Danesh Science and Technology University*
Dinajpur, Bangladesh
mfwahid26@gmail.com

Md. Fahim Shahriar
*Department of Electrical and Electronic Engineering*
*Hajee Mohammad Danesh Science and Technology University*
Dinajpur, Bangladesh
fahimshahriar6832@gmail.com

Md. Shohanur Islam Sobuj
*Department of Electrical and Electronic Engineering*
*Hajee Mohammad Danesh Science and Technology University*
Dinajpur, Bangladesh
shohanursobuj@gmail.com

*Abstract*—Bangla Handwritten Digit recognition is a significant step forward in the development of Bangla OCR. However, intricate shape, structural likeness and distinctive composition style of Bangla digits makes it relatively challenging to distinguish. Thus, in this paper, we benchmarked four rigorous classifiers to recognize Bangla Handwritten Digit: K-Nearest Neighbor (KNN), Support Vector Machine (SVM), Random Forest (RF), and Gradient-Boosted Decision Trees (GBDT) based on three handcrafted feature extraction techniques: Histogram of Oriented Gradients (HOG), Local Binary Pattern (LBP), and Gabor filter on four publicly available Bangla handwriting digits datasets: NumtaDB, CMARTdb, Ekush and BDRW. Here, handcrafted feature extraction methods are used to extract features from the dataset image, which are then utilized to train machine learning classifiers to identify Bangla handwritten digits. We further fine-tuned the hyperparameters of the classification algorithms in order to acquire the finest Bangla handwritten digits recognition performance from these algorithms, and among all the models we employed, the HOG features combined with SVM model (HOG+SVM) attained the best performance metrics across all datasets. The recognition accuracy of the HOG+SVM method on the NumtaDB, CMARTdb, Ekush and BDRW datasets reached 93.32%, 98.08%, 95.68% and 89.68%, respectively as well as we compared the model performance with recent state-of-art methods.

## I. Introduction

Bangla is the world's seventh most spoken mother tongue, with approximately 228 million native speakers, and the world's sixth most spoken language, with approximately 268 million total speakers (228 million native speakers and 37 million as second language speakers) [1]. In Bangladesh, it is the official language, while in India; it is the second most widely spoken language. Despite its widespread use, there has not been nearly as much research as there should have been to improve the computer suitability of the language. As a consequence, the Bangla language lacks a developed Optical Character Recognition (OCR) system. OCR, on the other hand, offers a wide range of uses, including autonomous driving, postal code automation, number plate recognition, document digitization, and so on [2]. Therefore, researchers are now focusing their efforts on developing Bangla OCR. Handwritten Character Recognition is a subset of OCR that focuses to recognize handwritten letters. Table I listed all of the digits in Bangla.

In the realm of digit recognition, a lot of work has been done on English handwritten digit recognition using handicraft feature extraction approaches; however there are just a few notable works in Bangla handwritten digit recognition using this technique. On the CMARTdb dataset, Saha et al. [2] employed LBP-based feature descriptors in conjunction with a linear kernel SVM to successfully detect Bangla basic characters, digits, and compound characters. They completed the task of recognition with a decent outcome. Hassan et al. [3] used the CMARTdb dataset with LBP to extract features. They utilized a KNN classifier to recognize handwritten Bangla numerals and acquired 96.7% accuracy. Khan et al. [4] used a sparse representation classifier to classify Bangla digits, and their method achieved 94% accuracy on the CMARTdb dataset. Choudhury et al. [5] suggested a feature extraction technique based on HOG and colour histograms, and they employed SVM as a classifier for Bangla digit identification. On the CMARTdb dataset, their suggested technique achieved 94% accuracy with the HOG feature and 98.05% accuracy with the HOG and color histogram features. A hybrid deep model with HOG features was proposed by Sharif et al. [6] where they have used two datasets: ISI numeral dataset and CMARTdb. The proposed model combines hand-crafted feature extraction with automatically learned features, achieving maximum accuracy of 99.02 % and 99.17 % on the ISI numeral dataset and CMARTdb dataset, respectively. In another work, a HOG features combined with SVM model is reported [7] which achieved 97.08% recognition accuracy on Mendeley data. Islam et al. [8] analyzed all publicly available Bangla handwritten digit datasets, including NumtaDB, CMARTdb, Ekush, and BDRW, as well as five robust algorithms: KNN, SVM, RF, MLP, and CNN, to benchmark them. On the Ekush dataset, the CNN model performs the best across all datasets, with 99.71% accuracy.

TABLE I. BANGLA DIGITS

| ০ | ১ | ২ | ৩ | ৪ |
|---|---|---|---|---|
| ৫ | ৬ | ৭ | ৮ | ৯ |

According to the above-mentioned literature on the Bangla handwritten digit recognition task, researchers usually investigate their model on a single dataset using handcrafted feature extraction techniques, whereas the authors of [8] used five classification algorithms and four publicly available datasets but did not use handcrafted feature extraction techniques. In this paper, we benchmarked four machine learning classifiers (KNN, SVM, RF, and GBDT) for Bangla handwritten digit recognition using three handcrafted feature extraction techniques (HOG, LBP, and Gabor) and four publicly available Bangla handwritten digit datasets (NumtaDB, CMARTdb, Ekush, and BDRW). We discovered that handcrafted features increased classification performance significantly, and that HOG features combined with an SVM model worked well on all datasets.

## II. MATERIALS AND METHODS

This section describes the overall methodology which is adopted for this research work. We have collected four publicly available well known Bangla handwritten digit dataset and employ three handcrafted feature extraction method to extracted feature from the dataset image. Then, these features are used to train four potential machine learning algorithm in order to classify Bangla handwritten digits. Fig. 1 represents the block diagram of the complete process.

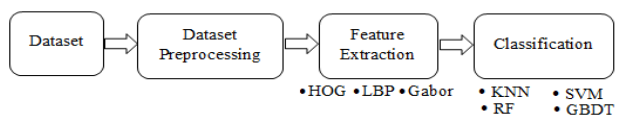

Fig. 1. Work flow of the research

### A. Data Collection

Data is an important component of machine learning. In order to evaluate the effectiveness of the handcrafted feature extraction technique and artificial intelligence classification algorithms, we have managed to collect four open sourced Bangla handwritten digit dataset namely NumtaDB [10], CMARTdb [3], Ekush [9] and BDRW [11]. Table II shows the total number of samples available in different datasets.

TABLE II.  DATASET DETAILS

| Dataset | Total Samples |
|---|---|
| NumtaDB | 85,596 |
| CMARTdb | 6,000 |
| Ekush | 30,688 |
| BDRW | 1,393 |

### B. Data Preprocessing

Machine learning algorithms rely heavily on data in a suitable format to perform adequately. Therefore, the collected data for this research work was pre-processed using image processing techniques in order to prevent misclassification and minimize computability. The images of the datasets include various forms of noise, such as varying sizes, gaussian noise and skew defects. To eliminate these noises, we first resized all the images to 28×28 pixels and then converted to grayscale images. Afterwards, Gaussian filters were used to minimize Gaussian noise which in turns reduces the risk of misclassification due to poor image quality. Finally, in the case of skew correction, if the image skews is positive, it will be rotated clockwise to correct it, otherwise, it will rotate anti-clockwise. Furthermore, we used the ".csv" version of the datasets to reduce computability time.

### C. Feature Extraction

Feature extraction is absolutely necessary in the field of machine learning for distinguishing images. Features are the peculiar signatures of a given image that describe it. Features are extracted to discriminate between the images. Various handcrafted algorithms like histogram of oriented gradients (HOG), local binary pattern (LBP), Scale-invariant feature transform (SIFT), Gabor are used to extract feature from images. In this paper, we used HOG, LBP, and Gabor feature extraction techniques for handwritten digit recognition.

*1) Histogram Oriented Gradients (HOG):* HOG is a feature descriptor used for human body detection that was first proposed by Dalal and Triggs [5]. It is widely used for object detection in computer vision and image processing. The HOG descriptor is divided the image into cells. Each cell contains a 1-d histogram of oriented gradient direction. In this paper, we used a [4 4] cell size that is grouped into a [2 2] block size. HOG characteristics obtained from a single image using cell size [2 2] and [8 8] are shown in Fig. 2. We have seen that in Fig. 2, the cell size [2 2] contains more shape information than the cell size [8 8]. We have used the [4 4] cell size for classification in this paper because it limits the number of dimensions and speeds up the training process. It also has enough information to visualize the shape of the digit.

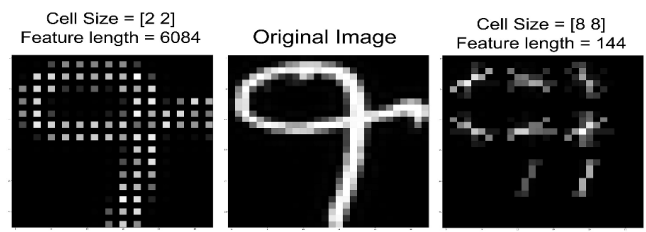

Fig. 2. HOG features of an Ekush dataset image

*2) Local Binary Pattern (LBP):* LBP is an efficient method used for texture extraction [3]. It calculates the local representation of the texture. It is a visual descriptor used in computer vision to categorize objects. It calculates the N×N neighbors of the center pixel value and converts the result to a binary number. LBP operates with different numbers of radius and neighbors. In this paper, we have set neighbors as 10, radius as 3 and default method to determine the pattern. Fig. 3 illustrates orginal image and the LBP-transformed image.

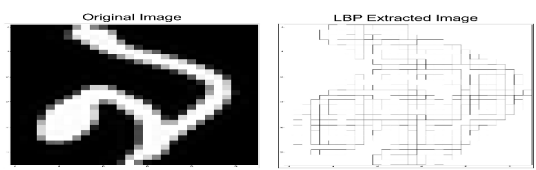

Fig. 3. LBP extracted feature for NumtaDB dataset

*3) Gabor Filter:* Gabor filter is a linear filter used in image processing for feature detection and texture analysis. It determines whether there is specific frequency information in the image with a specific orientation of a localized area around the region of analysis. The Gabor filter function extraction process begins with a 2D Gabor filter that is applied individually to each image. A Gabor filter oriented in a particular direction gives a strong response for locations of the target images that have structures in given direction. In this paper we used 2-D Gabor filter with frequency of 0.9 for texture and other parameters sets into default values.

## D. Machine Learning Classification Algorithms

In this paper, we utilize four robust classification algorithms namely KNN, SVM, RF and GBDT for classifying Bangla handwritten digits. We have fine-tuned all the classification algorithms in order to find optimal hyper-parameters for achieving benchmark performance.

*1) K-Nearest Neighbors (KNN):* K-nearest neighbor is a non-parametric machine learning algorithm for classification that is both simple and effective [3]. In KNN classification, an object is being assigned to a class that is most common among its k-nearest neighbors. Therefore, the best choice of hyper-parameter 'k' maximizes the classification performance. We have found that the best performing value of k is 5 for this work by employing hyper-parameter tuning. Here, we have used the 'Minkowski' distance metric and set the value of p=2 which is equivalent to the Euclidean metric.

*2) Support Vector Machine (SVM):* SVM is a supervised learning algorithm used for classification and regression analysis in machine learning. It tries to find a hyper-plane that maximizes the margin. For multiclass classification, it applies one-vs-all techniques. Various SVM kernels have been employed for classification but the RBF kernel shows pretty decent performance for the proposed classification task. The hyper-parameter, 'C' and ' 'gamma' values have also been fine-tuned to make the SVM robust.

*3) Random Forest (RF):* Random Forest is one of the most widely used bagging approaches, with its roots in the decision tree algorithm. It applies bagging technique with column sampling on top of it which reduces variance. It works well with large amounts of data with a reasonable amount of features. In this work, the hyper-parameter n_estimators was set to 200 and max_depth was set to 10.

*4) Gradient-Boosted Decision Trees (GBDT):* GBDT as an effective machine-learning technique gained popularity recently because of its high accuracy, fast training and prediction times, and small memory size. In the XGBOOST library, we have found n_estimator =200 and max_ depth=5 after hyper-parameter tuning for proposed classification task. It has also been noticed that the performance of GBDT has increased significantly as we used GBDT with row sampling and column sampling.

## III. RESULTS AND DISCUSSIONS

In order to accomplish the goal of Bangla Handwritten Digit Recognition, the performance of four classifiers (KNN, SVM, RF, GBDT) based on handcrafted features (HOG, LBP, Gabor) extracted from these datasets (NumtaDB, CMARTdb, Ekush, BDRW) was explored. The accuracy, precision, recall, and F1-score of the classifier were used to assess its performance. These evaluation metrics were calculated using the correctly and incorrectly classified classes from the confusion matrix. Furthermore, in order to train the classifier, 80% data from the dataset were used and the rest 20% of the data were used for testing purposes.

The overall performance of the four classifiers employed for Bangla Handwritten Digit Recognition on four different datasets based on handcrafted features is shown in Table III, Table IV, Table V and Table VI. It is observed from Table III and Table VI that both the classifiers KNN and GBDT shows higher accuracy on CMARTdb, Ekush, BDRW datasets using HOG features compared to LBP and Gabor features whereas in NumtaDB dataset both the classifier achieved highest accuracy using Gabor features. On the other hand, Table IV and Table V indicate that the accuracy of both SVM and RF classifiers employing HOG features was higher than LBP and Gabor features on all of the datasets studied. This was expected since HOG features preserve local pixel interactions and second order statistics, allowing it to achieve better recognition accuracy.

TABLE III. PERFORMANCE OF KNN

| Dataset | Feature | Accuracy (%) | Precision (%) | Recall (%) | F1-Score (%) |
|---|---|---|---|---|---|
| NumtaDB | HOG | 86.36 | 99.46 | 97.02 | 98.22 |
|  | LBP | 80.69 | 98.80 | 95.13 | 96.93 |
|  | Gabor | 87.52 | 99.46 | 98.48 | 98.97 |
| CMARTdb | HOG | 96.04 | 100 | 100 | 100 |
|  | LBP | 93.75 | 98.95 | 100 | 99.47 |
|  | Gabor | 95.10 | 100 | 100 | 100 |
| Ekush | HOG | 93.31 | 99.37 | 98.76 | 99.06 |
|  | LBP | 88.40 | 99.78 | 96.50 | 98.12 |
|  | Gabor | 88.40 | 99.78 | 96.50 | 98.12 |
| BDRW | HOG | 87.44 | 90.47 | 90.47 | 90.47 |
|  | LBP | 75.33 | 100 | 95.45 | 97.67 |
|  | Gabor | 82.51 | 100 | 96.15 | 98.03 |

TABLE IV. PERFORMANCE OF SVM

| Dataset | Feature | Accuracy (%) | Precision (%) | Recall (%) | F1-Score (%) |
|---|---|---|---|---|---|
| NumtaDB | HOG | 93.32 | 99.37 | 99.09 | 99.23 |
|  | LBP | 83.58 | 98.36 | 99.12 | 98.74 |
|  | Gabor | 75.34 | 95.04 | 99.03 | 96.99 |
| CMARTdb | HOG | 98.08 | 100 | 100 | 100 |
|  | LBP | 95.20 | 100 | 100 | 100 |
|  | Gabor | 95.62 | 100 | 100 | 100 |
| Ekush | HOG | 95.68 | 99.37 | 99.78 | 99.57 |
|  | LBP | 89.15 | 99.33 | 98.89 | 99.11 |
|  | Gabor | 89.15 | 99.33 | 98.99 | 99.11 |
| BDRW | HOG | 89.68 | 100 | 92.30 | 96.00 |
|  | LBP | 65.47 | 69.23 | 94.73 | 79.99 |
|  | Gabor | 82.51 | 95.83 | 92.00 | 93.87 |

TABLE V. PERFORMANCE OF RF

| Dataset | Feature | Accuracy (%) | Precision (%) | Recall (%) | F1-Score (%) |
|---|---|---|---|---|---|
| NumtaDB | HOG | 81.68 | 97.11 | 97.59 | 97.35 |
|  | LBP | 66.22 | 94.80 | 97.56 | 96.16 |
|  | Gabor | 78.77 | 98.11 | 97.24 | 97.67 |
| CMARTdb | HOG | 94.58 | 100 | 100 | 100 |
|  | LBP | 92.39 | 100 | 100 | 100 |
|  | Gabor | 94.37 | 100 | 100 | 100 |
| Ekush | HOG | 90.25 | 99.54 | 100 | 99.77 |
|  | LBP | 84.98 | 98.13 | 99.29 | 98.70 |
|  | Gabor | 92.68 | 99.35 | 99.57 | 99.46 |
| BDRW | HOG | 86.54 | 90.47 | 100 | 95.00 |
|  | LBP | 65.91 | 64.00 | 94.17 | 76.19 |
|  | Gabor | 78.00 | 95.45 | 91.30 | 93.33 |

The accuracy comparison of different classifier on NumtaDB, CMARTdb, Ekush, and BDRW datasets based on handcrafted features (HOG, LBP, and Gabor) are shown in Fig. 4. It is explored from the Fig. 4 that HOG+SVM model attained highest classification performance on all the dataset studied. The performance of the best performing models on different datasets for recognition of Bangla handwritten digit is listed in Table VII.

TABLE VI. PERFORMANCE OF GBDT

| Dataset | Feature | Accuracy (%) | Precision (%) | Recall (%) | F1-Score (%) |
|---|---|---|---|---|---|
| NumtaDB | HOG | 90.27 | 98.62 | 98.35 | 98.49 |
| NumtaDB | LBP | 87.20 | 99.15 | 99.06 | 99.10 |
| NumtaDB | Gabor | 90.74 | 99.26 | 98.8 | 99.03 |
| CMARTdb | HOG | 95.93 | 100 | 98.93 | 99.46 |
| CMARTdb | LBP | 94.48 | 100 | 98.92 | 99.45 |
| CMARTdb | Gabor | 94.79 | 100 | 100 | 100 |
| Ekush | HOG | 94.66 | 99.36 | 99.57 | 99.47 |
| Ekush | LBP | 90.83 | 99.33 | 98.45 | 98.89 |
| Ekush | Gabor | 92.68 | 99.35 | 99.57 | 99.46 |
| BDRW | HOG | 83.40 | 100 | 100 | 100 |
| BDRW | LBP | 65.00 | 84.21 | 94.11 | 88.11 |
| BDRW | Gabor | 70.40 | 95.00 | 90.47 | 92.68 |

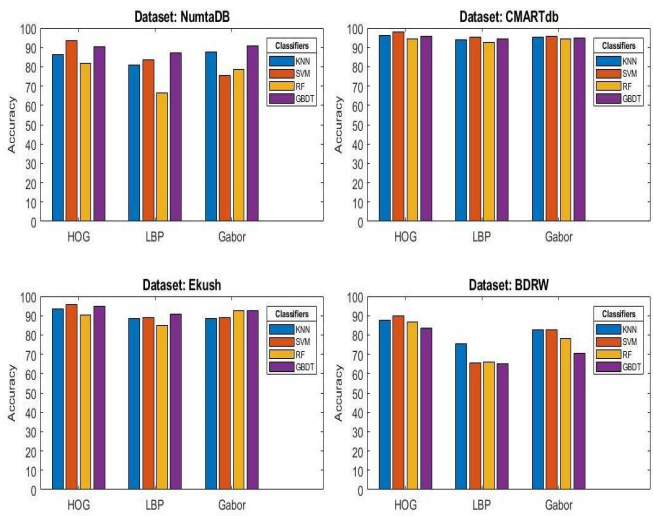

Fig. 4. Accuracy comparison of classifier on different datasets based on HOG, LBP and Gabor feature

TABLE VII. BEST PERFORMING MODELS

| Dataset | Model | Accuracy (%) | Precision (%) | Recall (%) | F1-Score (%) |
|---|---|---|---|---|---|
| NumtaDB | HOG+SVM | 93.32 | 99.37 | 99.09 | 99.23 |
| CMARTdb | HOG+SVM | 98.08 | 100 | 100 | 100 |
| Ekush | HOG+SVM | 95.68 | 99.37 | 99.78 | 99.57 |
| BDRW | HOG+SVM | 89.68 | 100 | 92.30 | 96.00 |

A comparative study is also conducted in order to demonstrate the effectiveness of handcrafted features on classification performance. Handcrafted features greatly improved the classification performance of Bangla Handwritten Digit recognition, as seen in Table VIII.

TABLE VIII. COMPARISON OF ACCURACY WITH AND WITHOUT FEATURE EXTRACTION

| Dataset | Accuracy without Feature Extraction (SVM) [8] | Accuracy with Feature Extraction (HOG+SVM) [Proposed] |
|---|---|---|
| NumtaDB | 78.00% | 93.32% |
| CMARTdb | 95.01% | 98.08% |
| Ekush | 90.72% | 95.68% |
| BDRW | 74.55% | 89.68% |

In order to assess its competence, the proposed HOG+SVM model is compared to contemporary state-of-the-art approaches. It is discovered that the proposed HOG+SVM model outperformed the reported models mentioned in Table XI.

TABLE IX. COMPARISON OF ACCURACY WITH RECENT WORKS

| Ref | Model | Dataset | Accuracy |
|---|---|---|---|
| [3] | LBP+KNN | CMARTdb | 96.70% |
| [5] | HOG+SVM | CMARTdb | 94.00% |
| [7] | HOG+SVM | Mendeley Data | 97.08% |
| Proposed | HOG+SVM | CMARTdb | 98.08% |

IV. CONCLUSION

This paper utilizes four different Bangla Handwritten digit datasets (NumtaDB, CMARTdb, Ekush, and BDRW) to demonstrate the impact of handcrafted features (HOG, LBP, Gabor) on classification performance of four different classifiers (KNN, SVM, RF, GBDT). We also fine-tuned the classification algorithms to select the most optimal hyperparameters. The handcrafted feature was discovered to have a significant impact on classification accuracy. The study reveals that HOG features combined with SVM (HOG+SVM) achieved the maximum accuracy on all of the datasets considered in this research work. The HOG+SVM approach attained recognition accuracy of 93.32%, 98.08%, 95.68% and 89.68% on NumtaDB, CMARTdb, Ekush, and BDRW datasets respectively. In future, deep learning based features, feature integration and feature selection techniques will be explored in order to boost the performance of Bangla Handwritten digit recognition techniques.